\documentclass{article}

\PassOptionsToPackage{square, numbers}{natbib}

\usepackage[preprint]{neurips_2026}

\usepackage{amsmath}
\usepackage{amsthm}
\usepackage{amssymb}
\usepackage{bm}
\usepackage{graphicx}
\usepackage{placeins}
\usepackage{booktabs}
\usepackage{pifont}
\usepackage{xcolor}
\usepackage{wrapfig}
\usepackage{enumitem}
\usepackage{url}

\newcommand{\X}{\bm{X}}

\newcommand{\I}{\bm{I}}

\newcommand{\cmark}{\textcolor{green!60!black}{\ding{51}}}
\newcommand{\xmark}{\textcolor{red}{\ding{55}}}

\theoremstyle{definition}
\newtheorem{definition}{Definition}
\newtheorem{example}{Example}

\title{Covariance Shrinkage via Stochastic Interpolation}

\author{
  Mathieu Chalvidal  \\
    Capital Fund Management\\
    23 Rue de l’Université, 75007 Paris\\
    \texttt{mathieu.chalvidal@cfm.com}
  \And
  Florentin Coeurdoux\\
    Capital Fund Management\\
    23 Rue de l’Université, 75007 Paris\\
  \texttt{florentin.coeurdoux@cfm.com}
  \And
    Eric Vanden-Eijnden\\
    Capital Fund Management\\
    23 Rue de l’Université, 75007 Paris\\
  \texttt{eric.vanden-eijnden@cfm.com}
}

\begin{document}

\maketitle

\begin{abstract}
We recast classical shrinkage of high-dimensional covariance estimators as empirical risk minimization over a parametric stochastic interpolant between a source and a target distribution. This formalism recovers known shrinkage estimators as special cases and reveals three distinct mechanisms for reducing statistical risk: \textbf{(i) Scheduling}: the interpolant schedule determines the class of admissible covariances, and hence the achievable risk; \textbf{(ii) Flow maps and couplings}: whereas naive constructions amount to assuming independence between the distributions, specific coupling structures (e.g., solutions of optimal transport problems can lower the empirical risk; moreover, non-linear flow maps realizing such couplings free the interpolant covariance from the eigenbasis of the empirical estimate, enabling eigenvector regularization; \textbf{(iii) Early stopping}: estimators defined by integrating a regressed vector field afford an additional bias–variance trade-off through approximation of the true interpolant distribution. We then propose a neural estimator of the interpolant, together with an upper bound on its quadratic risk in terms of the interpolant approximation error, and validate both on synthetic experiments. Finally, we apply the estimator to real neuroimaging data, demonstrating the regularization power this approach offers in practice.
\end{abstract}

\section{Introduction}

\textbf{Problem formulation} $\diamond$ Estimating the covariance matrix $\Sigma_\mu$ of a $D$-dimensional random vector $\smash{X \sim \mu}$ from $N$ samples is foundational across statistics and machine learning. When $D$ is non-negligible compared to $N$, the empirical sample covariance $\smash{\hat\Sigma}$ is high-variance and ill-conditioned: its spectrum spreads according to Marčenko--Pastur \citep{MP1967} and its eigenvectors absorb sampling noise. Classical \emph{shrinkage} \citep{LEDOIT2004365} regularizes $\smash{\hat\Sigma}$ by pulling its eigenvalues toward an anchor (identity, scalar-trace, factor model), trading bias for variance. Non-linear shrinkage refinements \citep{ledoit2020analytical, Lam2016, BUN20171} reshape the spectrum further but remain rotationally invariant: the empirical eigenvectors are preserved exactly, and any signal misaligned with $\smash{\hat\Sigma}$'s eigenbasis is unrecoverable.

\textbf{Main contributions} $\diamond$ We recast covariance shrinkage as empirical risk minimization over a parametric stochastic interpolant \citep{albergo2025stochastic} between an isotropic source $\mu_0$ and the empirical measure $\hat\mu$. This formalism unifies classical estimators as instances of a single construction and exposes three distinct axes of regularization:
\begin{itemize}[leftmargin=1em]
    \item \emph{Schedule.} The interpolant defines a two-parameter family of admissible covariances on $[0,1]^2$; the risk-minimizer generically lies in the interior, away from the linear and trace-preserving paths along which Ledoit--Wolf shrinkage is recovered.
    \item \emph{Coupling.} Optimal-transport and learned couplings reduce the achievable risk relative to the independent coupling. Non-linear flow maps additionally free the estimator from the empirical eigenbasis, lifting the rotational-invariance limitation that constrains all classical shrinkage estimators.
    \item \emph{Training budget.} For a neural velocity field $\hat v_t^\tau$, training to convergence collapses the estimator back to $\smash{\hat\Sigma}$. We observe that early-stopping improves the interpolant risk profile and calibrate the training budget directly from the data via either cross-validation or via a Stein-corrected risk surrogate that, unlike Stein--Haff, applies without rotational invariance.
\end{itemize}
We validate the estimator on synthetic Gaussian benchmarks across spectrum shapes and dimensionality ratios $q = D/N$, and on resting-state fMRI in the data-limited regime $q = 2$, where it outperforms Ledoit--Wolf and Wasserstein-2 shrinkage in terms of out-of-sample negative log-likelihood.

\begin{table}[!t]
  \centering
  \begin{tabular}{lllcc}
    \toprule
    \textbf{Interpolation type} & \textbf{Coupling} & \textbf{Solution} & \textbf{Spectral reg.} & \textbf{Eigenbasis reg.} \\
    \midrule
    Linear interp.  & Independent       & Analytical     & \cmark & \xmark \\
    %Linear SI         & Deterministic     & Linear     & \cmark & \xmark \\
    McCann interp.   & Optimal transport & Gradient flow ODE & \cmark & \cmark \\
    Stochastic interp.    & Deterministic     & Neural ODE & \cmark & \cmark \\
    \bottomrule
  \end{tabular}
  \vspace{0.15cm}
  \caption{Comparison of the interpolation processes studied in this paper, viewed as estimators between an isotropic anchor $\mu_0$ and the empirical measure $\hat\mu$. Different choices of coupling and velocity field yield different regularization effects; only the non-linear flow maps (rows 2--3) escape the rotational-invariance limitation of classical shrinkage.}
  \label{tab:model_comparison}
  \vspace{-0.5cm}
\end{table}

\section{Related work}
\textbf{Shrinkage methods} $\diamond$ 
James and Stein's realization \citep{james1961estimation} that multivariate least-square estimators are dominated by estimators that "shrink" parameters towards any target estimate paved the way for widespread regularization frameworks, ranging from classical Tikhonov-Ridge regularization \citep{ridge} to the lasso \citep{tibshirani1996regression} and various generalizations \citep{ElasticNet2005,Lasso2006,Robust2009,Concavegroupselection2015}. More specifically, shrinkage methods have been extensively developed to mitigate the high variance incurred by the empirical estimation of high-dimensional covariance matrices: \citep{LEDOIT2004365, ledoit2020analytical, BODNAR2016223} developed methods to select \textit{a priori} the optimal shrinkage intensity with respect to a target matrix.
Additionally, non-linear shrinkage of the matrix spectrum relying on random matrix theory have been proposed by \citep{BUN20171} with asymptotical optimality in the problem dimension D. Another popular framework for sparse precision matrix based on graphical models have been developed in \citep{cai2011constrainedl1minimizationapproach}. More recently, parameter shrinkage has been elegantly recast as the solution to a min-max optimization problem evaluated over a specific region of uncertainty \citep{yue2024geometric}. Importantly, in all these cases, the methods share a unifying structural assumption when the shrinkage target is simultaneously diagonalizable with the empirical estimate: they are rotationally invariant, meaning they only regularize the estimated spectrum while leaving the sample eigenvectors unchanged. 

\textbf{Data augmentation methods}: $\diamond$ Incorporating additional samples derived from the observed population have been shown to favorably improve empirical performance in numerous statistical inference applications \citep{vanDyk2001TheAO, MUMUNI2022100258} ranging from computer vision \citep{Khoshgoftaar2019} to graphs \citep{ding2022dataaugmentationdeepgraph} or text \citep{Khoshgoftaar2021}. The seminal work of \citep{bishop1995training} established an equivalence between isotropic Gaussian noise addition and Tikhonov regularization. Since then, some theoretical characterization of the beneficial effect of data augmentation techniques have been proposed in \citep{Re2019, Lee2020,Lin2022TheGT} through kernel and group theory, while augmentation techniques have been refined, in particular through modern generative modeling
\citep{Hansen2016, edwards2017} or automated augmentation strategies \citep{ratner2017, cubuk2019, NEURIPS2020_d85b63ef}. More recent work has however pointed out that iterative incorporation of synthetic data predictably hurt downstream regression performance due to the bias of an imperfect augmentation generator \citep{cetingoz2025syntheticdataportfoliosthrow, NEURIPS2024_53dbd7e3}. In limited data regime case, some efforts have also been dedicated to characterize the benefit of data augmentation, in particular for few-shot classification \citep{zhang2024few} or precision matrix estimation \citep{morisset2025,ZHANG2026106039}.

\section{Shrinkage as Stochastic Interpolation}

\textbf{Notations and conventions} $\diamond$ We denote by $D$ the dimensionality of $\X$ and by $N$ the number of observed realizations $\mathbb{X} = (X_1, \dots, X_N) \in \mathbb{R}^{D \times N}$. Throughout, we assume centered data, $\mathbb{E}_\mu[X] = 0$, so that the covariance of $X$ under any probability measure $\mu$ on $\mathbb{R}^D$ is the matrix-valued functional
\begin{equation}
    \Sigma_\mu := \mathbb{E}_\mu[XX^\top] = \int_{\mathbb{R}^D} x x^\top \, \mu(dx).
\end{equation}
The target of estimation is $\Sigma_\mu$, the covariance of the unknown data distribution $\mu$. The simplest estimator is the empirical sample covariance,
\begin{equation}
    \hat\Sigma := \Sigma_{\hat\mu} = \frac{1}{N} \sum_{n=1}^N X_n X_n^\top, \qquad \hat\mu := \frac{1}{N} \sum_{n=1}^N \delta_{X_n},
\end{equation}
which is unbiased but suffers from high variance when $D$ is comparable to or larger than $N$.

\textbf{Parametric construction and risk} $\diamond$ Our approach builds the estimator $\Sigma_\theta$ from a parametric random variable
\begin{equation}
    I_\theta = \mathcal{I}(X_0, X; \theta), \qquad \theta \in \Theta,
\end{equation}
where $X_0$ is an auxiliary variable drawn from a fixed source $\mu_0$ and $X$ is a target sample. Ideally we would draw $X \sim \mu$ from the true target, but in practice we only have access to the empirical measure $\hat\mu$ and must sample $X \sim \hat\mu$ instead. More generally the pair $(X_0, X)$ is drawn from a coupling $\nu \in \pi(\mu_0, \hat\mu)$ whose target-side marginal is $\hat\mu$ rather than $\mu$. Since $\hat\mu$ depends on the sample $\mathbb{X}$, the coupling $\nu$ and all objects derived from it are random in $\mathbb{X}$ --- a dependence we leave implicit in the notation. Letting $\mu_\theta$ denote the law of $I_\theta$ (i.e. the pushforward of $\nu$ under $\mathcal{I}(\cdot, \cdot; \theta)$) the induced covariance estimator
\begin{equation}
    \Sigma_\theta := \Sigma_{\mu_\theta} = \mathbb{E}_\nu\!\big[ I_\theta I_\theta^\top \big]
\end{equation}
is itself a random matrix, and we select $\theta$ to minimize its expected distance to the target,
\begin{equation}
    \theta^* \;\in\; \arg\min_{\theta \in \Theta} \mathcal{R}(\theta), \qquad \mathcal{R}(\theta) := \mathbb{E}_{\mathbb{X} \sim \mu^{\otimes N}}\!\big[\, d(\Sigma_\theta, \Sigma_\mu) \,\big],
    \label{eq:MINIMIZER}
\end{equation}
where $d$ is a matrix distance --- typically the squared Frobenius distance $d_F^2(A,B) = \|A - B\|_F^2$ between real-valued matrices or the squared Bures-Wasserstein distance between SPD matrices. The error in $\Sigma_\theta$ thus decomposes into two sources: a model error from approximating $\mu$ by the parametric family $\{\mu_\theta\}$, and a sampling error from substituting $\hat\mu$ for $\mu$ in the coupling. The various forms of regularization developed below trade these two sources against each other.

\subsection{Interpolant schedule}

A natural construction for $I_\theta$ is the stochastic interpolant of \citet{albergo2025stochastic}, parameterized by $\theta = (\alpha,\beta) \in [0,1]^2$ between a source distribution $\mu_0$ and the target $\mu$.

\begin{definition}[Stochastic interpolant]\label{def:SI}
Given a couple of random variables $(X_0, X)$ with joint law $\nu \in \pi(\mu_0, \hat\mu)$, the stochastic interpolant is
\begin{equation}
    I_\theta = \alpha\, X_0 + \beta\, X, \qquad \theta = (\alpha,\beta) \in [0,1]^2,
    \label{eq:StoInt}
\end{equation}
whose associated law $\mu_\theta$ is  the pushforward of $\nu$ under $(x_0, x) \mapsto \alpha x_0 + \beta x$. The endpoints $\theta = (1,0)$ and $\theta = (0,1)$ recover $\mu_0$ and $\hat\mu$ respectively.
\end{definition}

Already the simplest realization of this construction --- independent coupling against the empirical measure --- exposes the mechanics: the interpolant defines a two-parameter family of admissible covariances on $[0,1]^2$, and the risk minimum within this family generically lies in the interior, away from any one-dimensional path.

\begin{wrapfigure}{r}{0.55\textwidth}
    \centering
    \vspace{-0.7cm}
    \includegraphics[width=0.55\textwidth]{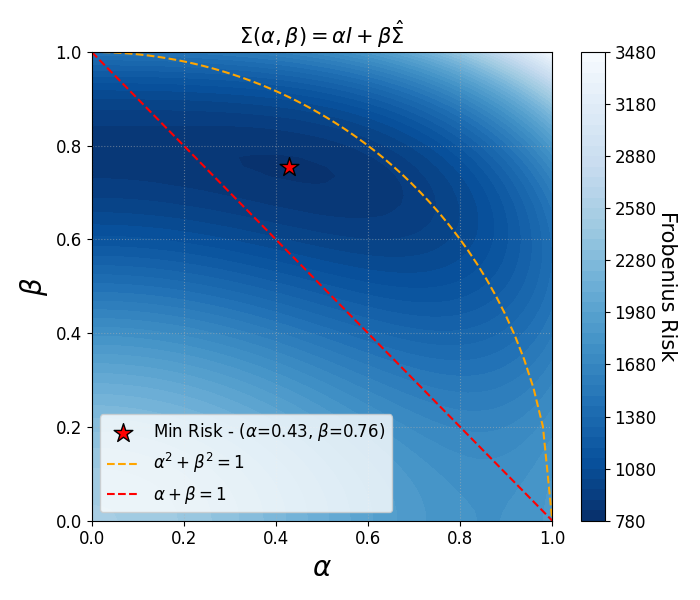}
    \vspace{-0.6cm}
    \caption{Frobenius risk surface $\mathcal{R}(\alpha,\beta)$ on $[0,1]^2$ for the interpolant between a 100-dimensional Gaussian target and an isotropic source, estimated from $N=100$ samples. The optimum lies neither on the linear path $\alpha+\beta=1$ nor on the trace-preserving path $\alpha^2+\beta^2=1$.}
    \label{fig:surface}
    %\vspace{-0.4cm}
\end{wrapfigure}

\begin{example}[Independent coupling]\label{ex:linear}
Take $\nu = \mu_0 \otimes \hat\mu$ with both $\mu_0$ and $\hat\mu$ centered. Independence and centering eliminate the cross-terms in $\mathbb{E}_\nu[I_\theta I_\theta^\top]$, and the interpolant covariance reduces to a quadratic form in the parameters,
\begin{equation}
    \Sigma_\theta = \alpha^2\, \Sigma_0 + \beta^2\, \hat\Sigma, \qquad \theta = (\alpha,\beta) \in [0,1]^2,
    \label{eq:LinearSI}
\end{equation}
with $\Sigma_0 := \Sigma_{\mu_0}$ and $\hat\Sigma := \Sigma_{\hat\mu}$. Two one-parameter restrictions are natural: the linear path $\alpha + \beta = 1$, and the trace-preserving path $\alpha^2 + \beta^2 = 1$. The latter, when $\Sigma_0 = I$ and $\mathrm{tr}(\Sigma_\mu) = D$, reduces \eqref{eq:LinearSI} to the convex combination $\Sigma_\theta = \lambda I + (1-\lambda)\hat\Sigma$ and recovers classical Ledoit--Wolf shrinkage \citep{LEDOIT2004365}. The Frobenius risk on the full square,
\begin{equation}
    \mathcal{R}^F(\alpha, \beta) = \mathbb{E}_{\mathbb{X}}\!\left[\, \big\| \alpha^2\, \Sigma_0 + \beta^2\, \hat\Sigma \;-\; \Sigma_\mu \big\|_F^2 \,\right],
    \label{eq:risk_explicit}
\end{equation}
admits a closed-form minimizer $(\alpha^*, \beta^*) \in [0,1]^2$. If $\hat\Sigma = \Sigma_\mu$ almost surely the minimizer collapses to $(0,1)$; any finite-sample deviation forces $\alpha^* > 0$, so a strictly shrunken estimator achieves lower risk irrespective of the anchor $\mu_0$ --- a finite-sample echo of Stein's inadmissibility result \citep{Stein}. Whenever $\Sigma_0$ is diagonal, however, the estimator preserves the eigenbasis of $\smash{\hat\Sigma}$ and is therefore \emph{rotationally invariant} \citep{BUN20171} --- a structural limitation we revisit in §3.2.
\end{example}

The independent coupling $\nu = \mu_0 \otimes \hat\mu$ of Example~\ref{ex:linear} was chosen for analytic tractability, but it sits at one extreme of the admissible set $\pi(\mu_0, \hat\mu)$ and the resulting estimator pays for that simplicity in variance. Enriching the family of couplings considered in \eqref{eq:MINIMIZER} can only lower the achievable risk, provided the new $\Sigma_\theta$ remains computable. 

\begin{example}[Gaussian optimal-transport coupling]\label{ex:OT}
Take both source and target Gaussian and centered, $\mu_0 = \mathcal{N}(0, \Sigma_0)$ and $\hat\mu = \mathcal{N}(0, \smash{\hat\Sigma})$. The Monge--Kantorovich map under the Wasserstein-2 distance is then affine, $X_1 = T(X_0) = A X_0$, with
\begin{equation}
    A = \Sigma_0^{-1/2}\big(\Sigma_0^{1/2}\, \hat\Sigma\, \Sigma_0^{1/2}\big)^{1/2} \Sigma_0^{-1/2},
    \label{eq:OT_map}
\end{equation}
and the displacement geodesic $I_t = (1-t)\, X_0 + t\, A X_0$ realizes the interpolant of Definition~\ref{def:SI} along the linear path $\theta_t = (1-t, t)$ under this OT-induced coupling. Specializing to an isotropic source $\Sigma_0 = \sigma^2 I$ collapses $A$ to $\smash{\sigma^{-1} \hat\Sigma^{1/2}}$ and yields the closed-form covariance
\begin{equation}
    \Sigma_t^{\mathrm{OT}} = (1-t)^2 \sigma^2 I + 2t(1-t)\, \sigma\, \hat\Sigma^{1/2} + t^2\, \hat\Sigma.
    \label{eq:OT_cov}
\end{equation}
At the spectral level, writing $\hat\lambda_i$ for the eigenvalues of $\hat\Sigma$,
\begin{equation}
    \lambda_i^{\mathrm{OT}}(t) = \big((1-t)\,\sigma + t\, \sqrt{\hat\lambda_i}\,\big)^2,
    \label{eq:OT_spectrum}
\end{equation}
so the OT-induced shrinkage interpolates eigenvalues \emph{linearly in standard deviation} rather than in variance. Compared to the independent coupling of Example~\ref{ex:linear}, which gives $\smash{\lambda_i^{\mathrm{lin}}(t) = (1-t)^2\sigma^2 + t^2 \hat\lambda_i}$, each eigenvalue under \eqref{eq:OT_spectrum} now moves monotonically from $\sigma^2$ to $\smash{\hat\lambda_i}$ along $t \in [0,1]$: large eigenvalues ($\smash{\hat\lambda_i > \sigma^2}$) are not depressed at intermediate $t$, and small eigenvalues ($\smash{\hat\lambda_i < \sigma^2}$) descend more gradually. Both effects are favorable in the finite-sample regime where $\smash{\hat\Sigma}$ generically over-estimates large eigenvalues and under-estimates small ones \citep{MP1967}. We note that $A$ shares its eigenbasis with $\smash{\hat\Sigma}$, so $\Sigma_t^{\mathrm{OT}}$ is diagonal in that basis and the estimator remains rotationally invariant --- a limitation we lift in §3.3 by parameterizing the velocity field with a neural network.
\end{example}

\subsection{Coupling via flow maps}
We consider next ordinary differential equations of the form 
\begin{equation}
    \dot X_t = v_t(X_t), \qquad X_0 \sim \mu_0,
    \label{eq:flow_ode}
\end{equation}
which transports samples from $\mu_0$ at $t=0$ to a terminal distribution $\mu_1$ at $t=1$, with intermediate marginals $\mu_t$ satisfying the continuity equation
\begin{equation}
    \partial_t \mu_t + \nabla \cdot (\mu_t\, v_t) = 0, \quad \mu_{t=0}=\mu_0
    \label{eq:cont_eq}
\end{equation}
The pair $(X_0, X_1)$ defines a deterministic coupling in $\pi(\mu_0, \mu_1)$, and we focus on velocity fields for which $\mu_1 = \hat\mu$, exactly or in approximation. Two natural constructions arise:

\textbf{Gradient flows of an optimal-transport potential.} $\diamond$  Following the Benamou--Brenier formulation \citep{benamou2000}, the velocity field minimizing $\smash{\int_0^1 \mathbb{E}_{\mu_t}[\|v_t\|^2]\,dt}$ subject to \eqref{eq:cont_eq} and the boundary conditions $\mu_0, \mu_1 = \hat\mu$ generates the Wasserstein-2 optimal transport plan as its endpoint coupling. In practice the flow is realized by descending the gradient of an entropic OT distance with respect to particle positions transported from $\mu_0$ to $\hat\mu$.

\textbf{Conditional flow matching.} $\diamond$ Restricting Definition~\ref{def:SI} to a one-parameter path $\theta_t = (\alpha_t, \beta_t)$ with $\alpha_0 = \beta_1 = 1$ and $\alpha_1 = \beta_0 = 0$  conditional flow matching \citep{lipman2023flow, albergo2023building} regresses a parametric velocity field $v_\phi$ against the conditional expectation
\begin{equation}
    v_t(x) = \mathbb{E}_\nu\!\big[\, \dot I_t \,\big|\, I_t = x \,\big],\quad I_t = I_{\theta_t} = \alpha_t X_0 + \beta_t X
    \label{eq:cfm}
\end{equation}
where $\nu \in \pi(\mu_0, \hat\mu)$ is a chosen base coupling. The velocity $v_t(x)$ can be learned via minimization of the loss
\begin{equation}
    L[\hat v] = \mathbb{E}_{\nu,t\sim U([0,1])}\!\big[|\hat v_t(I_t) - \dot I_t |^2\big],
    \label{eq:cfm_target}
\end{equation}

over a rich parametric class of $\hat v_t(x)$. The minimizer is nonlinear in $x$ in general, so the flow can encode higher-order structure of $\hat\mu$ that the independent coupling of Example~\ref{ex:linear} cannot capture.

\begin{figure}[h!]
  \begin{center}
\includegraphics[width=\textwidth]{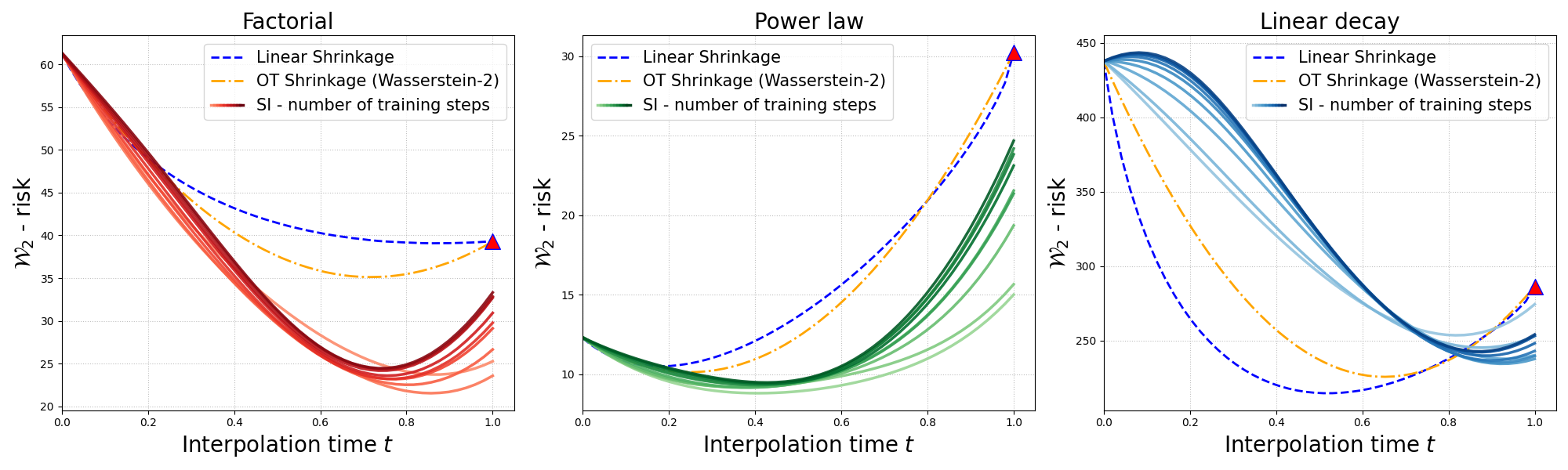}
  \end{center}
  \vspace{-0.3cm}
  \caption{Risk profiles under the Bures-Wasserstein distance of several interpolant constructions for three different Gaussian distributions described in Section \ref{sec:experiments_control}. The empirical estimate is pictured as a red triangle. Color darkness indicate vector field training progress ranging linearly from 1K to 50K gradient steps. OT shrinkage is constructed analytically from the empirical covariance. Depending on the underlying distribution, the SI estimator might largely reduce the risk.} 
  \label{fig:non_linear_panel}
\end{figure}

\begin{figure}[h!]
  \begin{center}
\includegraphics[width=\textwidth]{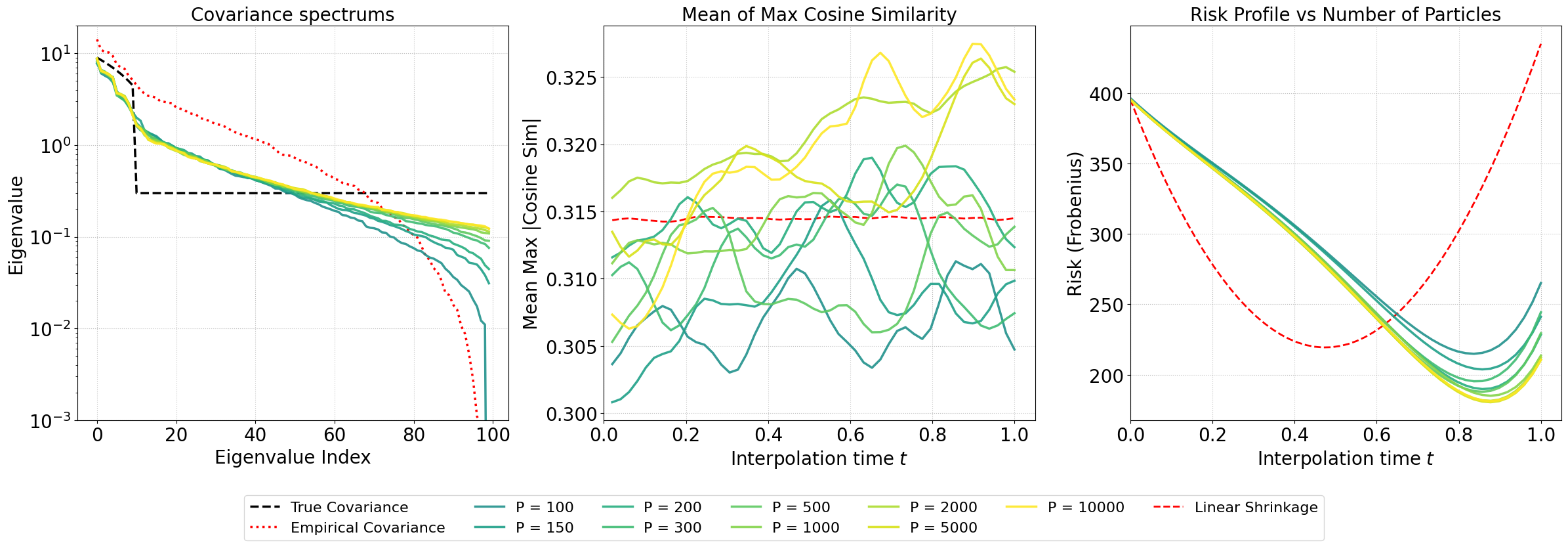}
\vspace{-0.7cm}
  \end{center}
  \caption{Evolution of the covariance estimator with the number of samples considered for the Monte-Carlo estimation. Finite sample effects affecting the empirical estimate get largely attenuated in the spectrum tail (\textbf{left}). In parallel, the alignment of the true versus estimated eigendirections increases (\textbf{middle}). These two effects contribute to the reduction of risk.(\textbf{right})} 
  \label{fig:mc_decay}
\end{figure}

\textit{Monte Carlo over flow trajectories:} A practical benefit of building the coupling through a flow map is that once $v_t$ is fixed, sampling from $\mu_t$ at any $t$ amounts to integrating \eqref{eq:flow_ode} forward, which can be done over arbitrarily many independent trajectories. The estimator $\Sigma_\theta$ is then evaluated by Monte Carlo estimation over $M \gg N$ samples, replacing the $\mathcal{O}(N^{-1/2})$ sampling error inherited from $\hat\mu$ by an integration error of order $\smash{M^{-1/2}}$ that can be driven below the model bias at modest cost. Figure~\ref{fig:mc_decay} quantifies this effect.

\subsection{Early stopping}

When the velocity field in \eqref{eq:flow_ode} is parameterized by a neural network $\hat v_t$ trained with a conditional-flow-matching loss \citep{lipman2023flow}, the training budget $\tau$ controls how closely $\hat v_t^\tau$ approximates the regression target \eqref{eq:cfm_target}. For a sufficiently expressive network and $\tau \to \infty$, the trained flow pushes $\mu_0$ exactly onto $\hat\mu$ and the induced covariance estimator collapses to the empirical sample covariance: $\smash{\Sigma_\theta \to \hat\Sigma}$. The model memorizes the sampling noise of $\hat\mu$ and the regularization purchased by the schedule and the coupling is undone. At finite capacity the hypothesis class of $\hat v_t$ imposes additional implicit regularization on the converged flow; this effect is real but orthogonal to our analysis and we do not study it here.

Stopping training before this fixed point is reached prevents the collapse. At intermediate $\tau$, the trained field $\hat v_t^\tau$ generates a smoothed terminal distribution whose covariance lies between that of the isotropic source and $\smash{\hat\Sigma}$, trading the bias of an under-trained flow against the sampling variance of $\smash{\hat\Sigma}$. The training budget therefore enters the risk \eqref{eq:MINIMIZER} as a third regularization axis, alongside the schedule $(\alpha_t, \beta_t)$ of §3.1 and the coupling $\nu$ of §3.2. The next subsection shows how to estimate this risk from the data and select $\tau^*$ accordingly.

\begin{figure}[h!]
  \begin{center}
\includegraphics[width=\textwidth]{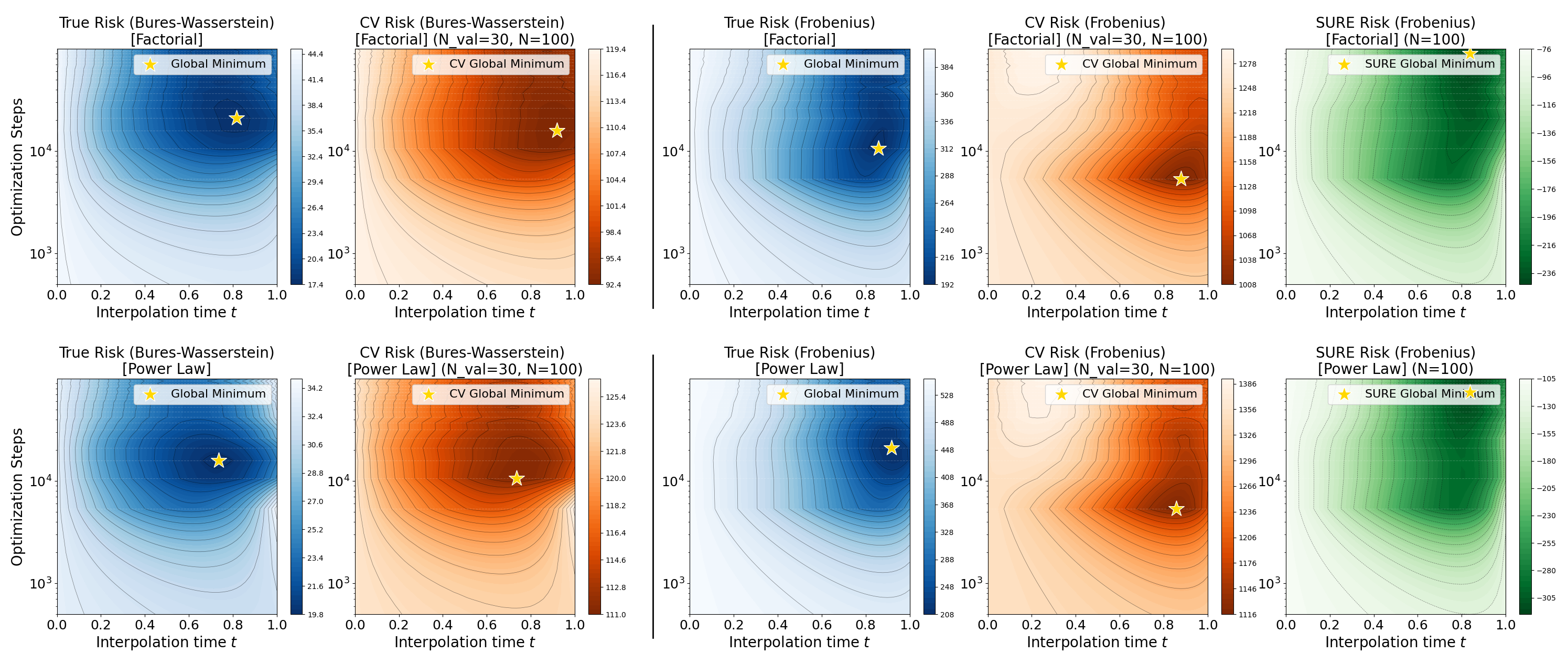}
  \end{center}
  \vspace{-0.3cm}
  \caption{Comparison of true versus estimated risk profiles for the factorial and power law models considered in section \ref{sec:experiments_control} for Bures-Wasserstein (\textbf{left}) and Frobenius (\textbf{right}) distances. For the CV estimation, a held out set of $N_{val} =30$ samples is used} 
\end{figure}

\subsection{Risk estimation}
\label{sec:theory_sure}

To deploy the estimator we need a data-driven proxy for the risk $\mathcal{R}(\theta)$ in \eqref{eq:MINIMIZER}. We propose two complementary methods.

\textbf{Cross-validation.} Split the sample into a training subset $\mathbb{X}_{\mathrm{tr}}$ used to construct $\Sigma_\theta$ and a held-out subset $\mathbb{X}_{\mathrm{val}}$ used to evaluate it, and approximate $\mathcal{R}(\theta)$ by the held-out distance $\smash{d\big(\Sigma_\theta(\mathbb{X}_{\mathrm{tr}}),\, \hat\Sigma_{\mathrm{val}}\big)}$, where $\smash{\hat\Sigma_{\mathrm{val}}}$ is the empirical covariance on the validation fold. Although the split further reduces the data available for fitting --- a real concern in the small-sample regime --- we show empirically that a modest validation fold is sufficient to obtain stable risk estimates even at high dimension.

\textbf{SURE correction.} Specializing $\mathcal{R}(\theta)$ to the squared Frobenius distance,
\begin{equation}
    \mathcal{R}^F(\theta) = \mathbb{E}_{\mathbb{X}}\!\big[\|\Sigma_\theta\|_F^2\big] \;-\; 2\, \mathbb{E}_{\mathbb{X}}\!\big[\mathrm{Tr}(\Sigma_\theta\, \Sigma_\mu)\big] + \|\Sigma_\mu\|_F^2,
    \label{eq:R_F_decomposition}
\end{equation}
only the cross term depends on the unobservable $\Sigma_\mu$. Naively replacing it by $\smash{\hat\Sigma}$ underestimates the risk because $\Sigma_\theta$ was built from the same data, leading to an optimistic bias. The standard correction comes from a multivariate Stein identity: assuming $X_n \sim \mathcal{N}(0, \Sigma_\mu)$ and applying Stein's lemma componentwise to each $X_n$ yields (proof in Appendix~\ref{app:sure})
\begin{equation}
    \mathbb{E}_{\mathbb{X}}\!\big[\mathrm{Tr}(\Sigma_\theta\, \Sigma_\mu)\big] = \mathbb{E}_{\mathbb{X}}\!\big[\mathrm{Tr}(\Sigma_\theta\, \hat\Sigma)\big] \;-\; \frac{1}{N} \sum_{n=1}^N \mathbb{E}_{\mathbb{X}}\!\big[\mathrm{Tr}\big(\Sigma_\mu \cdot \mathrm{jvp}_{X_n}(\Sigma_\theta,\, X_n)\big)\big],
    \label{eq:stein_identity}
\end{equation}
where $\mathrm{jvp}_{X_n}(\Sigma_\theta, X_n) \in \mathbb{R}^{D \times D}$ denotes the Jacobian of $\Sigma_\theta$ with respect to $X_n$ contracted with $X_n$ itself. Substituting back into \eqref{eq:R_F_decomposition} and dropping the $\theta$-independent constant $\|\Sigma_\mu\|_F^2$ gives the risk surrogate
\begin{equation}
    \tilde{\mathcal{R}}(\theta) = \mathbb{E}_{\mathbb{X}}\!\big[\|\Sigma_\theta\|_F^2\big] \;-\; 2\, \mathbb{E}_{\mathbb{X}}\!\big[\mathrm{Tr}(\Sigma_\theta\, \hat\Sigma)\big] + \frac{2}{N} \sum_{n=1}^N \mathbb{E}_{\mathbb{X}}\!\big[\mathrm{Tr}\big(\Sigma_\mu \cdot \mathrm{jvp}_{X_n}(\Sigma_\theta,\, X_n)\big)\big].
    \label{eq:SURE}
\end{equation}
The correction term in \eqref{eq:SURE} still contains $\Sigma_\mu$, so the surrogate is not directly observable. In practice we plug in $\smash{\hat\Sigma}$ for $\Sigma_\mu$ inside the correction, yielding an asymptotically unbiased risk estimator that --- in contrast to Stein--Haff applied to rotationally-invariant shrinkage --- requires no special structure on $\Sigma_\theta$ and is therefore compatible with the neural interpolant of §3.3.

% \subsection{Error bounds}

% Finally, we provide an error bound for the risk of this estimator against the true interpolant statistic $\mathcal{E}^*_t  = \mathbb{E}_{\nu}[\varphi(I_t)]$. In the case of Frobenius norm, we can show that:
% \begin{equation}
%     \vert\vert \mathcal{E}^{\theta,t}_\varphi - \mathcal{E}^*_t\vert \vert_F \leq  \frac{M \mathcal{L}_{SI}(t)}{2L}\int_{0}^{t}\sqrt{e^{2Ls}-1}ds \leq \frac{M \mathcal{L}_{SI}(t)}{2L}(e^{Lt}-1) 
% \end{equation}

\section{Numerical experiments}

\subsection{Controlled experiment}
\label{sec:experiments_control}

In this section we empirically compare the risk of several interpolant estimators across different settings where we control the data generative process, thus having access to the ground truth covariance. We vary the underlying distribution, the amount of available data and the dimensionality of the considered estimation problems. We also consider two risk definitions arising from either the Bures-Wasserstein distance between SPD matrices or the Frobenius norm between real-valued matrices. 

\textbf{Experiment 1} $\diamond$ We first consider 100-dimensional Gaussian distributions with different spectral properties. For each of the three settings considered below, we sample 100 points such that we place our setting in $q = 1$. Additionally, to make the estimation problem closer to real-case scenarios, we further inject a sparse ($p=0.1$) Gaussian noise $\epsilon$ with a small scale $(\sigma_{\epsilon}=0.1)$ to our observations.
\begin{itemize}[leftmargin=2em]
    \item \textit{Factorial model}: This case is prevalent in signal processing and machine learning. The distribution has most of its variance concentrated on a small number of dimensions within the ambient space while the variance of other dimensions is set to a nominal level $c=0.5$.
    \item \textit{Power law spectrum}: The covariance spectrum is decaying with a power law $s\approx t^{-1}$.  
    \item \textit{Linear decay spectrum} : Contrary to the two previous models, the variance is spread over all dimensions with a linear decay. This case is most adversarial to small data population estimation since the variance is more evenly spread over each dimension.
\end{itemize}
For each distribution, we train a 2-layer multi-layer perceptron with ReLU activations as our neural interpolant with schedule $\alpha_t + \beta_t = 1$ for 50.000 steps on the full batch and compute the covariance by sampling $n=20.000$ trajectories.  Results are presented in figure \ref{fig:non_linear_panel}. We show that the risk can be notably reduced compared to vanilla shrinkage methods in both settings where the prior is closer or farther from the ground truth than the empirical estimate, while increasing as training progress past a certain training budget (data overfitting).

\begin{figure}[h!]
  \begin{center}
\includegraphics[width=\textwidth]{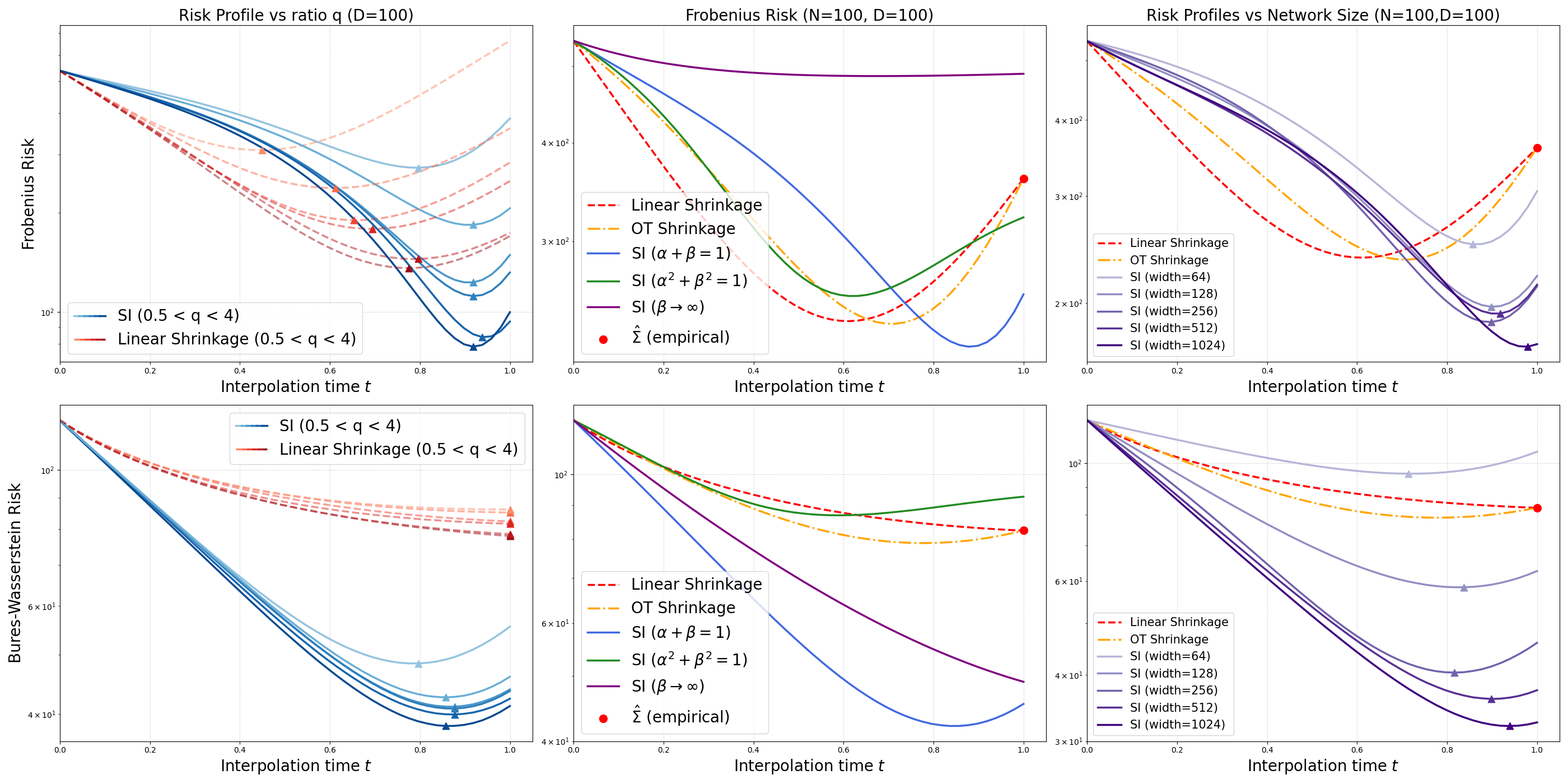}
  \end{center}
  \vspace{-0.1cm}
  \caption{Risk profiles for covariance estimation under the two considered distances varied across different specifications for an elliptic 100-dimensional Gaussian distribution corrupted with sparse random noise (see experiment 1). The covariance path learned by the neural interpolant reach a lower minimal error than the linear interpolation, providing a different bias-variance trade-off that depends on ratio $q=D/N$ (\textbf{left}), the interpolant definition (\textbf{middle}) and the network capacity (\textbf{right}). 
  % This regularization gap is particularly strong for covariance with low conditionning number and decrease gracefully with better conditionning (\textbf{right})
  } 
  \label{fig:panel_profiles}
  \vspace{-0cm}
\end{figure}

\textbf{Experiment 2} $\diamond$ We investigate next how the neural interpolant estimator behaves as a function of the data ratio $q$, the interpolant schedule and the model capacity. Even in very limited data regime, the regularization effect of the interpolation over linear shrinkage can remain, and is dependent on the schedule definition, while restricting the model dimensionality compared to the distribution increases the risk uniformly over the interpolant trajectory (see figure \ref{fig:panel_profiles}).

\subsection{Real world application}
\label{sec:experiments_fmri}

We validate our approach on a real world application of neuroimaging: Functional Magnetic Resonance Imaging (fMRI) refers to imaging methods able to capture spatial and temporal changes in brain metabolism that are induced by cognitive state changes or unregulated processes in the resting brain, through variation of oxygen concentration in blood flows. fMRI is being used as a biomarker for disease detection, to monitor therapy in patients, or for studying pharmacological efficacy. However, fMRI data natively operates in a highly adverse statistical regime. The number of parcellated brain regions ($D$) is typically large, while the number of temporal observations ($N$) per scanning session is limited due to scanner costs and subject motion constraints. In this $D \approx N$ or $D > N$ regime, the empirical sample covariance matrix is notoriously ill-conditioned or singular, severely overfitting the in-sample noise. Consequently, regularization is mandatory. 
% While classical linear techniques, such as Ledoit-Wolf (LW) shrinkage, provide well-conditioned estimators by pulling eigenvalues toward a global mean, they rigidly preserve the empirical eigenbasis. Brain networks, however, exhibit complex underlying structural couplings dictated by physical white-matter tracts. By utilizing non-linear, path-dependent regularization, our SI estimator respects the local second-order data geometry, allowing the estimator to deform away from the noisy empirical eigenbasis along the probability flow.

\subsubsection{Out-of-Sample Statistical Risk Assessment}
To empirically validate both our estimator's generalization capability and the efficacy of our proposed Stein's Unbiased Risk Estimate (SURE) early-stopping criterion, we conduct an out-of-sample risk evaluation on the ABIDE dataset \citep{di2014autism}. 

\textbf{Experimental Setup} $\diamond$
We utilize pre-extracted region-of-interest time series parcellated into $D = 200$ regions from the Craddock atlas \citep{craddock2012}. Each region's time series is independently standardized to zero mean and unit variance. To simulate the extreme data-limited regime, each subject's time series $X \in \mathbb{R}^{N \times D}$ is split longitudinally into a training set $X_{\text{train}}$ of length $N_{\text{train}} = 100$, and a held-out test set $X_{\text{test}}$ of length $N_{\text{test}} = 100$. This enforces a dimensionality ratio of $q = D/N = 2.0$, ensuring the empirical training covariance $\smash{\hat{\Sigma}_{\text{train}}}$ is singular.

\textbf{Estimator Calibration} $\diamond$
For each subject, we construct the SI estimator path $\Sigma_t$ initialized at the isotropic source distribution $\Sigma_0 = I$. The optimal integration time $t^*$ is selected strictly using the training data by minimizing the theoretical SURE penalty derived in Section~\ref{sec:theory_sure}, specifically utilizing batched Jacobian-Vector Products (JVPs) over $X_{\text{train}}$. The final calibrated estimator is $\smash{\hat{\Sigma}_{\text{FM}} = \Sigma_{t^*}}$. As benchmark, we consider the Ledoit-Wolf linearly shrunk covariance $\smash{\hat{\Sigma}_{\text{LW}}}$ and the Wasserstein-2 OT regularization $\smash{\hat{\Sigma}_{\text{OT}}}$.

\paragraph{Evaluation Metric}
The performance of the estimators is measured using the out-of-sample Gaussian Negative Log-Likelihood (NLL) on the unseen test acquisitions. Importantly, to bypass the rank-deficiency of the empirical test covariance $\hat{\Sigma}_{\text{test}}$, the average NLL is evaluated using the stable trace formulation:
\begin{equation}
    \mathcal{L}(\Sigma) \propto \frac{1}{2} \left( \log \det(\Sigma) + \text{Tr}(\Sigma^{-1} \hat{\Sigma}_{\text{test}}) \right)
\end{equation}
where $\hat{\Sigma}_{\text{test}} = \frac{1}{N_{\text{test}}} X_{\text{test}}^T X_{\text{test}}$. Note that this formulation only requires the inversion of the well-conditioned, regularized estimators.

\subsection{Results}
The mean out-of-sample Negative Log-Likelihood across all subjects is reported in Table~\ref{tab:fmri_nll}. The unregularized sample covariance cannot be evaluated due to its singularity ($\det(\hat{\Sigma}_{\text{train}}) = 0$).

\begin{table}[h]
\centering
\begin{tabular}{lrr}
\toprule
\textbf{Estimator} & \textbf{Out-of-Sample NLL} ($\downarrow$) & \textbf{Computation time} ($\downarrow$) \\
\midrule
Sample Covariance ($\hat{\Sigma}_{\text{train}}$) & N/A (Singular) & $\bm{\sim 100\mu s}$ \\
Ledoit-Wolf Linear Shrinkage ($\hat{\Sigma}_{\text{LW}}$) & 51.68 $\pm$ 42.81 & $\sim 10ms$ \\
Wasserstein-2 Shrinkage ($\hat{\Sigma}_{\text{OT}}$) & 40.89 $\pm$ 24.44 & $\sim 30 s$ \\
\midrule
\textbf{SI Shrinkage (Ours)} & \textbf{30.96 $\pm$ 15.64} & $\sim 7 s$ \\
\bottomrule
\end{tabular}
\vspace{0.25cm}
\caption{Out-of-Sample Negative Log-Likelihood on held-out fMRI acquisitions ($D=200$, $N_{\text{train}}=100$, $N_{\text{test}}=100$). Lower is better. NLL is reported as Mean $\pm$ Standard Deviation across subjects.}
\label{tab:fmri_nll}
\end{table}

As shown in Table~\ref{tab:fmri_nll}, our SI shrinkage Estimator significantly outperforms both classical linear shrinkage and optimal transport regularized shrinkage. By successfully leveraging the SURE stopping criterion, the SI estimator halts the probability flow at an optimal geometry that generalizes better to unseen fMRI acquisitions, proving the practical utility of non-linear covariance interpolation in high-dimensional biological settings.

\section{Discussion}

By reframing covariance shrinkage as the minimization of a parametric empirical risk over a continuous-time flow, we unite classical algebraic methods—such as linear and optimal transport shrinkage—under a single continuous framework. Beyond serving as a unifying theory, this formalism reveals new pathways for reducing statistical risk by viewing the interpolant schedule, the coupling structure, and the integration time as distinct axes of regularization. A primary conceptual advantage of this framework is its ability to bypass the rotational invariance constraint that strictly limits traditional shrinkage techniques. Through non-linear flow maps parameterized by neural networks, our approach dynamically regularizes the empirical eigenvectors. This enables the estimator to exploit the local second-order geometry of the data by performing an annealed, path-dependent shrinkage that gracefully adjusts to the underlying data manifold. Furthermore, casting the estimation as the integration of a vector field naturally introduces the integration trajectory as an explicit control over the bias-variance trade-off. We operationalized this insight using a bias penalization, providing a fully observable, data-driven criterion to halt the continuous flow at an optimal geometry before overfitting occurs. The practical efficacy of this approach was validated in the highly adverse, data-limited regime of neuro-imaging, where the non-linear interpolant captured the covariance structure significantly better than established robust baselines.\\

\textbf{Limitations and Future Work}
While the neural interpolant demonstrates powerful geometric regularization, parameterizing the velocity field with a neural network intrinsically carries a higher computational burden than deriving closed-form analytical estimators. Future research will explore more computationally efficient parameterizations of the flow maps to accelerate the integration process. Additionally, the stochastic interpolation formalism naturally lends itself to the estimation of higher-order statistical moments, presenting an exciting frontier for building robust, non-parametric estimators in complex geometric spaces.

\newpage
\bibliographystyle{unsrtnat}
\bibliography{refs}

\newpage 
\appendix

\section{Theoretical Details}

\subsection{Effect of conditional couplings on Frobenius risk} 
Let us keep the same construction for $\mathcal{I}(\mathbb{X},\theta)$ as in Definition 1, but consider more complex couplings $\nu$ between $\X_0$ and $\X$. In the general case, it is possible to express our interpolant estimator as :
\begin{equation}
        \mathbb{E}_{\rho_{\theta}}[\varphi(\I(\theta))] = \alpha^2\mathbb{E}_{\rho_0}[\varphi(\X_0)] + \beta^2\mathbb{E}_{\rho}[\varphi(\X)] +  2\alpha\beta\mathbb{E}_{\nu}[\varphi(\X_0,\X)]
\end{equation}

where $\varphi(\X_0,\X)$ corresponds to the cross-correlation statistic. Here, setting $\smash{\beta = (1-\alpha)}$ and observing that $\varphi(\X_0,\X) = \varphi(\X_0) + \varphi(\X) - \varphi(\X-\X_0)$ , yields 
% $\mathbb{E}_{\rho_{\theta}}[\varphi(\I(\theta))] = \mathcal{E}_{\nu}^{\varphi,\perp}(t) -  \alpha(1-\alpha)\mathbb{E}_{\nu}[\varphi(\X_0-\X)]$. Thus, recomputing equation \eqref{eq:risk_explicit} for
the associated risk $\mathcal{R}_{\rho}^{\varphi}(t)$:
\begin{equation}
    \mathcal{R}_{\varphi}(\alpha) = \mathcal{R}_{\varphi}^{\perp}(\alpha) - 2\alpha(1-\alpha)\mathbb{E}_{\mathbb{X}}\big[\langle \mathcal{E}_{\nu}^{\varphi,\perp}(\alpha) -  \mathbb{E}_{\rho}[\varphi(\X)], \mathbb{E}_{\nu}[\varphi(\X_0-\X)] \rangle_F\big] + \mathcal{O}(\alpha^2(1-\alpha)^2)
    %\mathbb{E}_{\mathbb{X}}\big[\mathbb{E}_{\nu}[\varphi(\X_0-\X)]\big]
\end{equation}
where $\mathcal{R}^{\perp}_{\varphi}(\alpha)$ corresponds to the independent coupling case. We see that a coupling $\nu$ where the difference between the  error between the statistic under the distribution $\X_0$ and the empirical estimator is anticorrelated, favorably reduces the risk ($\alpha, \beta \geq 0 )$ and describe a different landscape for the risk.  \\

\subsection{Stein unbiased risk estimation}
\label{app:sure}

Recall the squared Frobenius risk of the estimator $\Sigma_t$,
\begin{equation}
  \label{eq:sure-decomp}
  \mathcal{R}^F(t)
  = \mathbb{E}_{\mathbf{X}}\!\left[ \|\Sigma_t - \Sigma_\mu\|_F^2 \right]
  = \underbrace{\mathbb{E}_{\mathbf{X}}\!\left[ \|\Sigma_t\|_F^2 \right]}_{\text{observable}}
  \;-\; 2\,\mathbb{E}_{\mathbf{X}}\!\left[ \operatorname{Tr}(\Sigma_t \Sigma_\mu) \right]
  \;+\; \underbrace{\|\Sigma_\mu\|_F^2}_{\text{constant}},
\end{equation}
where $\Sigma_\mu$ denotes the unknown target covariance and $\hat\Sigma = \tfrac{1}{N}\mathbf{X}\mathbf{X}^\top$ the sample covariance. The first term is computable from the estimator alone, and the third is independent of $t$ and may be dropped for optimization. The middle term is the only obstacle: it involves the unknown $\Sigma_\mu$, and because $\Sigma_t$ is itself constructed from $\mathbf{X}$, the naive substitution $\Sigma_\mu \!\to\! \hat\Sigma$ is optimistically biased.

To quantify this bias, expand the naive plug-in using the cyclic property of the trace,
\begin{equation}
  \label{eq:sure-plugin}
  \mathbb{E}_{\mathbf{X}}\!\left[ \operatorname{Tr}(\Sigma_t \hat\Sigma) \right]
  = \frac{1}{N}\sum_{i=1}^N \mathbb{E}_{\mathbf{X}}\!\left[ x_i^\top \Sigma_t\, x_i \right],
\end{equation}
which evaluates $\Sigma_t$ against the very samples used to construct it. The correction is supplied by the multivariate Stein identity: for $x \sim \mathcal{N}(0,\Sigma_\mu)$ and any  differentiable $f:\mathbb{R}^D\!\to\!\mathbb{R}^D$,
\begin{equation}
  \label{eq:stein}
  \mathbb{E}\!\left[ x^\top f(x) \right] = \mathbb{E}\!\left[ \operatorname{Tr}\!\big( \Sigma_\mu\, \nabla_x f(x) \big) \right].
\end{equation}
We apply \eqref{eq:stein} with $f(x_i) = \Sigma_t\, x_i$. Since $\Sigma_t$ depends on $x_i$, the product rule gives
\begin{equation}
  \label{eq:jvp}
  \nabla_{x_i}\!\big( \Sigma_t\, x_i \big) = \Sigma_t + \operatorname{jvp}_{x_i}(\Sigma_t, x_i),
  \qquad
  \operatorname{jvp}_{x_i}(\Sigma_t, x_i) := \big( \nabla_{x_i} \Sigma_t \big)[x_i] \in \mathbb{R}^{D\times D},
\end{equation}
where the second term is the directional derivative of the matrix $\Sigma_t$ contracted with $x_i$, i.e.\ a Jacobian--vector product. Substituting \eqref{eq:jvp} into \eqref{eq:stein} and averaging over the $N$ samples yields the identity
\begin{equation}
  \label{eq:sure-identity}
  \mathbb{E}_{\mathbf{X}}\!\left[ \operatorname{Tr}(\Sigma_\mu \Sigma_t) \right]
  = \mathbb{E}_{\mathbf{X}}\!\left[ \operatorname{Tr}(\Sigma_t \hat\Sigma) \right]
  - \frac{1}{N}\sum_{i=1}^N \mathbb{E}_{\mathbf{X}}\!\left[ \operatorname{Tr}\!\big( \Sigma_\mu \cdot \operatorname{jvp}_{x_i}(\Sigma_t, x_i) \big) \right].
\end{equation}
The first term on the right recovers the naive plug-in; the second is precisely the optimism bias it incurs. Inserting \eqref{eq:sure-identity} into \eqref{eq:sure-decomp} and discarding the constant $\|\Sigma_\mu\|_F^2$ gives the risk surrogate
\begin{equation}
  \label{eq:sure-surrogate}
  \widetilde{\mathcal{R}}(t)
  = \mathbb{E}_{\mathbf{X}}\!\left[ \|\Sigma_t\|_F^2 \right]
  - 2\,\mathbb{E}_{\mathbf{X}}\!\left[ \operatorname{Tr}(\Sigma_t \hat\Sigma) \right]
  + \frac{2}{N}\sum_{i=1}^N \mathbb{E}_{\mathbf{X}}\!\left[ \operatorname{Tr}\!\big( \Sigma_\mu \cdot \operatorname{jvp}_{x_i}(\Sigma_t, x_i) \big) \right].
\end{equation}
The penalty in \eqref{eq:sure-surrogate} still contains the unknown $\Sigma_\mu$, but only inside an $\mathcal{O}(1/N)$ term; replacing it by a high-quality empirical proxy therefore leaves the estimator asymptotically unbiased. Taking the proxy to be $\hat\Sigma$ recovers the standard SURE penalty, whereas taking the Ledoit--Wolf estimate $\hat\Sigma_{\mathrm{LW}}$ yields a lower-variance hybrid penalty in the small-sample regime. Finally, the sum of per-sample Jacobian--vector products equals a single batched product over $\mathbf{X}$, so the penalty deployed on the observed sample reads
\begin{equation}
  \label{eq:sure-penalty}
  \mathrm{Penalty}(t) = \frac{2}{N}\, \operatorname{Tr}\!\big( \hat\Sigma_{\mathrm{prox}} \cdot \operatorname{jvp}_{\mathbf{X}}(\Sigma_t, \mathbf{X}) \big),
  \qquad \hat\Sigma_{\mathrm{prox}} \in \{\, \hat\Sigma,\ \hat\Sigma_{\mathrm{LW}} \,\}.
\end{equation}

\section{Frobenius Error Bound for the Covariance Estimator}
\label{sec:weak_error_bound}

In this section, we derive a theoretical upper bound for the estimation error of our continuous-time covariance estimator. Rather than bounding the pointwise distance between the true and approximated trajectories (which typically yields loose bounds due to Grönwall's inequality), we directly bound the \textit{weak error}. This approach leverages the regularity of the true flow to measure how local vector field approximations compound into the final statistical estimator.

Let $\mu_0$ and $\hat{\mu}$ denote the initial and target distributions, coupled via $\nu \in \pi(\mu_0, \hat{\mu})$. The exact interpolant $I_t \in \mathbb{R}^D$ is governed by the true velocity field $v_t(x)$ via the ordinary differential equation (ODE) $dI_t = v_t(I_t)dt$. 
We define the exact flow map $\Phi_{t,s}: \mathbb{R}^D \to \mathbb{R}^D$, which transports a point from time $s$ to time $t$ along the true vector field, such that $\Phi_{s,s}(x) = x$ and $\frac{\partial}{\partial \tau} \Phi_{\tau, s}(x) = v_\tau(\Phi_{\tau, s}(x))$. Our covariance estimator relies on integrating a neural velocity field $v^\theta_t(x) \approx v_t(x)$ to produce an approximate trajectory $I^\theta_t$, satisfying $dI^\theta_t = v^\theta_t(I^\theta_t)dt$. The true and approximated terminal covariance matrices are respectively given by:
\begin{equation}
    \Sigma(t) = \mathbb{E}_{\nu}[I_t I_t^\top], \quad \text{and} \quad \Sigma_{\theta}(t) = \mathbb{E}_{\nu}[I^\theta_t (I^\theta_t)^\top]
\end{equation}
where $I_0 = I^\theta_0$ almost surely. Our objective is to bound the Frobenius risk $\vert\vert \Sigma_{\theta}(t) - \Sigma(t) \vert\vert_F$.

To bypass trajectory-wise error accumulation, we introduce a matrix-valued \textit{value function} $u(s, x) : [0, t] \times \mathbb{R}^D \to \mathbb{R}^{D \times D}$. For a given terminal time $t$ and spatial coordinate $x$, we define:
\begin{equation}
    u(s, x) = \Phi_{t,s}(x) \Phi_{t,s}(x)^\top
\end{equation}
The function $u(s, x)$ represents the terminal outer product obtained if we follow the \textit{true} flow starting from point $x$ at time $s$. By definition, evaluating this function at the boundaries yields:
\begin{equation}
    u(t, x) = xx^\top \quad \text{and} \quad u(0, I_0) = \Phi_{t,0}(I_0)\Phi_{t,0}(I_0)^\top = I_t I_t^\top \label{eq:boundary_u}
\end{equation}
Because $\Phi_{t,s}(x)$ represents a backward mapping from the perspective of time $s$, $u(s, x)$ satisfies the fundamental transport equation (the deterministic backward Kolmogorov equation):
\begin{equation}
    \partial_s u(s, x) + D_x u(s, x)[v_s(x)] = 0 \label{eq:pde}
\end{equation}
where $D_x u(s, x)[\cdot]$ denotes the directional derivative of $u$ with respect to the spatial variable $x$, acting as a linear operator from $\mathbb{R}^D$ to $\mathbb{R}^{D \times D}$.We can express the difference between the approximated and true terminal statistics by evaluating the value function along the approximated path $I^\theta_s$. Using the boundary conditions from Equation \eqref{eq:boundary_u}, we have:
\begin{equation}
    I^\theta_t (I^\theta_t)^\top - I_t I_t^\top = u(t, I^\theta_t) - u(0, I^\theta_0)
\end{equation}
Applying the fundamental theorem of calculus, we rewrite this difference as the integral of its total time derivative:
\begin{align}
    u(t, I^\theta_t) - u(0, I^\theta_0) &= \int_0^t \frac{d}{ds} u(s, I^\theta_s) ds \nonumber \\
    &= \int_0^t \Big( \partial_s u(s, I^\theta_s) + D_x u(s, I^\theta_s) \big[ \dot{I}^\theta_s \big] \Big) ds \nonumber \\
    &= \int_0^t \Big( \partial_s u(s, I^\theta_s) + D_x u(s, I^\theta_s) \big[ v^\theta_s(I^\theta_s) \big] \Big) ds
\end{align}
Substituting the transport equation identity $\partial_s u(s, I^\theta_s) = - D_x u(s, I^\theta_s)[v_s(I^\theta_s)]$ from Equation \eqref{eq:pde} into the integral, the partial derivative $\partial_s u$ elegantly cancels out, leaving only the residual of the vector fields:
\begin{equation}
    I^\theta_t (I^\theta_t)^\top - I_t I_t^\top = \int_0^t D_x u(s, I^\theta_s) \big[ v^\theta_s(I^\theta_s) - v_s(I^\theta_s) \big] ds \label{eq:defect}
\end{equation}
This formula acts as a non-linear variation of constants, mapping the local truncation error of the velocity field directly to the terminal covariance error, weighted by the spatial sensitivity $D_x u$ of the true flow. To bound Equation \eqref{eq:defect}, we must bound the operator norm of $D_x u(s, x)$. By applying the product rule to $u(s, x) = \Phi_{t,s}(x) \Phi_{t,s}(x)^\top$, the directional derivative applied to a perturbation vector $w \in \mathbb{R}^D$ is:
\begin{equation}
    D_x u(s, x)[w] = \big(J_{t,s}(x)w\big) \Phi_{t,s}(x)^\top + \Phi_{t,s}(x) \big(J_{t,s}(x)w\big)^\top
\end{equation}
where $J_{t,s}(x) = \frac{\partial}{\partial x} \Phi_{t,s}(x) \in \mathbb{R}^{D \times D}$ is the Jacobian matrix of the true flow. Taking the Frobenius norm and applying sub-multiplicativity yields:
\begin{equation}
    \vert\vert D_x u(s, x)[w] \vert\vert_F \leq 2 \vert\vert J_{t,s}(x) w \vert\vert_2 \vert\vert \Phi_{t,s}(x) \vert\vert_2 \leq 2 \vert\vert J_{t,s}(x) \vert\vert_{\text{op}} \vert\vert \Phi_{t,s}(x) \vert\vert_2 \vert\vert w \vert\vert_2
\end{equation}

Assuming the true vector field generates a well-behaved flow whose Jacobian norm is bounded by a constant $C_J(s) > 0$, the operator norm is bounded by $2 C_J(s) \vert\vert \Phi_{t,s}(x) \vert\vert_2$. We now take the expectation over the coupling $\nu$ and apply the Frobenius norm to Equation \eqref{eq:defect}. Passing the norm inside the integral and applying the Cauchy-Schwarz inequality over the expectation yields:
\begin{align}
    \vert\vert \Sigma_{\theta}(t) - \Sigma(t) \vert\vert_F &\leq \int_0^t \mathbb{E}_{\nu} \Big[ \vert\vert D_x u(s, I^\theta_s) \big[ v^\theta_s(I^\theta_s) - v_s(I^\theta_s) \big] \vert\vert_F \Big] ds \nonumber \\
    &\leq 2 \int_0^t C_J(s) \mathbb{E}_{\nu} \Big[ \vert\vert \Phi_{t,s}(I^\theta_s) \vert\vert_2 \vert\vert v^\theta_s(I^\theta_s) - v_s(I^\theta_s) \vert\vert_2 \Big] ds \nonumber \\
    &\leq 2 \int_0^t C_J(s) \sqrt{ \mathbb{E}_{\nu} \big[ \vert\vert \Phi_{t,s}(I^\theta_s) \vert\vert_2^2 \big] } \sqrt{ \mathbb{E}_{\nu} \big[ \vert\vert v^\theta_s(I^\theta_s) - v_s(I^\theta_s) \vert\vert_2^2 \big] } ds
\end{align}
Assuming bounded second moments for the flow paths, the term $\sqrt{ \mathbb{E}_{\nu} [ \vert\vert \Phi_{t,s}(I^\theta_s) \vert\vert_2^2 ] }$ is bounded by a spatial constant $R_t$. Defining $\kappa(s) = 2 R_t C_J(s)$, we arrive at our final bound:
\begin{equation}
    \vert\vert \Sigma_{\theta}(t) - \Sigma(t) \vert\vert_F \leq \int_0^t \kappa(s) \sqrt{ \mathbb{E}_{\nu} \Big[ \vert\vert v^\theta_s(I^\theta_s) - v_s(I^\theta_s) \vert\vert_2^2 \Big] } ds \label{eq:final_weak_bound}
\end{equation}
Equation \eqref{eq:final_weak_bound} demonstrates that the Frobenius risk of the covariance estimator scales linearly with the $L^2$ error of the neural velocity field. By bounding the weak error through the adjoint variable $u(s, x)$, we effectively avoid the exponential compounding associated with trajectory distance bounds. Instead, the error amplification is strictly governed by $\kappa(s)$, which relies purely on the structural regularity and stability (the Jacobian) of the \textit{true} target flow.

\section{Experimental Details}

\subsection{Evaluation Metrics}
We evaluate the estimated covariance matrices $\hat{\Sigma}$ against the ground-truth $\Sigma_{\text{true}}$ using the following metrics:
\begin{itemize}
    \item \textbf{Matrix distance:} The distance between the estimated covariance $\hat{\Sigma}$ and the true covariance $\Sigma_{\text{true}}$ is quantified using matrix distances. Specifically, we consider the following two metrics:
    \begin{itemize}
        \item \textbf{Frobenius Norm}, the standard entry-wise distance metric defined as:
        \[
        d_{\text{F}}^2(\hat{\Sigma},\Sigma_{\text{true}})= \sqrt{\text{Tr}\left((\hat{\Sigma} - \Sigma_{\text{true}})^2\right)}
        \]
        \item \textbf{Bures-Wasserstein Distance} defined between SPD matrices as the 2-Wasserstein distance between centered Gaussian distributions with such covariance matrices, defined as:
        \[
        d_{\text{BW}}^2(\hat{\Sigma}, \Sigma_{\text{true}}) = \text{Tr}(\hat{\Sigma}) + \text{Tr}(\Sigma_{\text{true}}) - 2\text{Tr}\left(\left((\hat{\Sigma})^{1/2} \Sigma_{\text{true}} (\hat{\Sigma})^{1/2}\right)^{1/2}\right)
        \]
    \end{itemize}
    \item \textbf{Spectral Alignment (Cosine Similarity):} We compute the maximum absolute cosine similarity between the eigenvectors of $\smash{\hat{\Sigma}}$ and the eigenvectors corresponding to the largest eigenvalues of $\Sigma_{\text{true}}$. The cumulative sum of these similarities, ordered by rank, provides a measure of how well the principal eigenspace is recovered.
\end{itemize}

\subsection{Synthetic experiments}

\subsubsection{Synthetic Data Generation Process}

To systematically evaluate the performance of our estimators, we employ various synthetic data generation processes that allow us to control the true underlying covariance structure and the nature of the observation noise. The data generation generally follows a two-step procedure: first, generating "clean" samples from a specified true covariance matrix $\Sigma_{\text{true}}$, and second, corrupting these samples with observation noise.

\textbf{Clean Data Generation Models}
We explore several structural archetypes for the true covariance matrix $\Sigma_{\text{true}} \in \mathbb{R}^{D \times D}$:

\begin{itemize}
    \item \textbf{Factorial Data (Spiked Covariance):} This model simulates a scenario with a few strong latent factors embedded in isotropic noise. The true eigenvalues $\lambda_i$ are set to a baseline noise level for most dimensions, while the first $k$ eigenvalues (the "spikes") decrease linearly from a specified signal strength. A random orthogonal matrix $Q$ is obtained via the QR decomposition of a standard Gaussian matrix. The true covariance is $\Sigma_{\text{true}} = Q \operatorname{diag}(\lambda_1, \dots, \lambda_D) Q^T$. Clean data $X_{\text{clean}} \in \mathbb{R}^{N \times D}$ is generated as $X_{\text{clean}} = Z \operatorname{diag}(\sqrt{\lambda_1}, \dots, \sqrt{\lambda_D}) Q^T$, where $Z_{ij} \sim \mathcal{N}(0, 1)$.
    
    \item \textbf{Power-Law Spectrum:} This model captures scenarios where the explained variance decays smoothly, as often seen in natural signals. The true covariance is diagonal, $\Sigma_{\text{true}} = \operatorname{diag}(\lambda_1, \dots, \lambda_D)$, with eigenvalues following a power-law decay: $\lambda_i = c \cdot i^{-\alpha} + \epsilon$, where $c$ is a scaling constant, $\alpha$ dictates the decay rate, and $\epsilon$ is a nominal noise floor. Clean data is sampled as $\smash{X_{\text{clean}} = Z \Sigma_{\text{true}}^{1/2}}$ for $\smash{Z_{ij} \sim \mathcal{N}(0, 1)}$.
    
    \item \textbf{Linear decay Spectrum:} This model represents scenarios where the variance is spread more homogeneously across the $D$ dimensions according to a uniform distribution with range [a,b] where b=$3\sigma_{\text{noise}}$ and a=$2\sigma_{\text{noise}}$ (see below)
\end{itemize}

\textbf{Observation Noise Models} To make our experiment more realistic, we additionally corrupt the clean data $X_{\text{clean}}$ to produce the observed empirical data $X_{\text{emp}}$. We considered in our experiments three distinct types of additive noise:

\begin{itemize}
    \item \textbf{Sparse Outliers:} A fraction of the data entries (determined by a corruption rate $p$) is replaced by extreme outliers. A binary mask $M$ is sampled where $M_{ij} \sim \text{Bernoulli}(p)$. The observed data is $X_{\text{emp}} = X_{\text{clean}} + M \odot O$, where $O_{ij} \sim \mathcal{N}(0,  \sigma_{\text{noise}}^2)$ and $\sigma_{\text{noise}}$ is the base noise level.
    
    \item \textbf{Heavy-Tailed Noise:} Dense noise is added to all entries, sampled from a Student's t-distribution with 2 degrees of freedom, scaled by $\sigma_{\text{noise}}$. This simulates data with heavy tails and frequent moderate outliers.
    
    \item \textbf{Heteroskedastic Noise:} Feature-dependent Gaussian noise is added, where the noise variance varies across dimensions. A set of standard deviations is linearly spaced between $0.1$ and $2\sigma_{\text{noise}}$ and then randomly permuted across the $D$ dimensions. Noise is sampled accordingly for each feature.
\end{itemize}

For the noise perturbation settings that we considered, we empirically observed minimal difference across the three settings with respect to the regularization effect of the learned interpolant, and report only the first case (sparse outliers). We emphasize that more completely characterizing the interpolant estimator properties as a function of signal-to-noise ratio might be valuable but is beyond this paper scope.

\subsubsection{Effect of Particle Count on Covariance Estimation} 

In this section, we detail the experimental setup used to investigate the effect of the number of simulated particles on the quality of the covariance estimator derived from the neural vector field in Figure 3. The experiment compares the recovered covariance spectrum, cosine similarity of eigenvectors, and risk profiles across various particle counts $P \in \{100, 150, 200, 300, 500, 1000, 2000, 5000, 10000\}$.

\textbf{Data Generation:} We generate synthetic high-dimensional data following the factor model. The dimensionality of the features is set to $D = 100$, and the number of observations is $N = 100$.

\textbf{Model Architecture and Training:}
A neural approximation of the vector field $v_\theta(x, t)$ is trained to transport an anchor Gaussian distribution $\mathcal{N}(0, \sigma_0^2 I)$ (with $\sigma_0 = 1.0$) to the empirical data distribution. A simple linear schedule is utilized for the interpolation process. We employ an MLP architecture with a hidden dimension of $256$. The model is trained using Adam with a learning rate of $10^{-4}$ for a total of $15,000$ epochs. 

\textbf{ODE Integration and Covariance Estimation:}
To evaluate the learned neural covariance at any intermediate time $t \in [0, 1]$, we sample $P$ initial particles $\smash{X_0^{(i)} \sim \mathcal{N}(0, \sigma_0^2 I)}$. The particles are pushed forward through the learned vector field $v_\theta$ using the Euler integration scheme. The time interval $t \in [0, 1]$ is discretized into $n_{\text{steps}} = 50$ evenly spaced steps ($\Delta t = 1/49$). The update rule is given by:
\[
X_{t + \Delta t}^{(i)} = X_t^{(i)} + v_\theta(X_t^{(i)}, t) \Delta t \quad \forall i \in \{1, \dots, P\}
\]
At each time step $t$, the estimated covariance matrix is computed as the empirical covariance of the current particle population:
\[
\Sigma_t^{(P)} = \frac{1}{P-1} \sum_{i=1}^P (X_t^{(i)} - \bar{X}_t)(X_t^{(i)} - \bar{X}_t)^\top
\]

\subsection{Shrinkage Interpolant Profiles}

 The experiments explored the impact of varying the data size ($N$), different interpolation schedules, and the neural network's hidden dimension on the estimation risk, measured by both Frobenius (top row) and Bures-Wasserstein distances (bottom row).

\paragraph{Experimental Setup} 
All experiments utilized a factorial data generation process for the true covariance $\Sigma_{\text{true}}$ with dimension $D=100$ and a signal strength of 10.0. Observed data was generated by adding sparse observation noise (noise level 0.3, corruption rate 0.1) to the clean data. The interpolation process started from an isotropic prior with $\sigma_0=1.0$.

\textbf{Panel 1: Impact of Data-to-Dimension Ratio (varying $N$)} The left column of Figure \ref{fig:panel_profiles} illustrates the true risk profiles for varying data-to-dimension ratios $q = N/D$, with $D=100$. The $N$ values explored range from $50$ to $300$, corresponding to $q$ ratios of $0.5$ to $3.0$. For each $N$, the SI method was trained with a fixed hidden dimension of 256. SI performance is compared against a Linear Shrinkage baseline. The color intensity indicates the $q$-ratio. Both risk measures consistently show that the SI method is able to find lower risk estimators across a range of $N$ values compared to the baseline, with minimum risk points generally observed at intermediate interpolation times $t$.

\textbf{Panel 2: Comparison of Interpolation Schedules} The middle column focuses on a fixed data size $N=100$ ($q=1.0$), comparing the performance of the SI method under different interpolation schedules: 'linear' ($\alpha+ \beta = 1$), 'vp' (variance-preserving, $\alpha^2 + \beta^2 = 1$), and 've' (variance-exploding, $\beta \rightarrow \infty$). These are benchmarked against traditional Linear Shrinkage and Optimal Transport (OT) Shrinkage. The results indicate that the 'linear' schedule consistently achieves the lowest risk for SI among the tested schedules, outperforming the analytical baselines. 

\textbf{Panel 3: Impact of Neural Network Capacity} The right column investigates the influence of the neural network's hidden dimension on the SI performance, keeping $N=100$ and $D=100$ fixed and retaining the network weights that achieve lower validation risk over the optimization trajectory. Hidden dimensions ranging from 64 to 1024 were tested. It is observed that increasing the hidden dimension generally leads to lower minimum risk values for the SI estimator. The SI method, particularly with larger hidden dimensions, robustly outperforms the Linear and OT shrinkage baselines.

\subsection{fMRI acquisition covariance estimation}

This section details the steps taken to acquire and prepare the neuroimaging data for covariance estimation.

\textbf{Data Acquisition:} The ABIDE PCP dataset, containing resting-state functional magnetic resonance imaging (fMRI) data, was utilized for this study. Specifically, pre-extracted time series data, processed with the CC200 atlas (resulting in 200 brain regions), were fetched using the \texttt{nilearn} library. The data was downloaded to a specified Google Drive directory for persistent storage and accessibility:

\begin{itemize}
    \item \textbf{Library:} \texttt{nilearn.datasets.fetch\_abide\_pcp}
    \item \textbf{Atlas:} CC200 (yielding \(D=200\) regions of interest)
    \item \textbf{Derivatives:} \texttt{\texttt{'rois\_cc200'}} (pre-extracted 2D arrays)
\end{itemize}

Upon fetching, the dataset object provides access to the individual subject's time series files. The first subject's time series, for example, had a shape of \((196, 200)\), indicating 196 time points and 200 brain regions.

\textbf{Data Filtering:} To ensure data quality and consistency, a filtering step was applied to the acquired time series. Only those subject time series \(i\) with more than 100 time points (i.e., \(i.\texttt{shape[0]} > 100\)) were retained for further analysis. This resulted in a subset of \(846\) subjects from the initial \(871\) subjects.

\textbf{Data Splitting:} For each subject's Z-scored time series \(X_{\text{zscored}}\), the data was split into training and testing sets. The first 100 time points were used for training \((X_{\text{train}})\) and the remaining time points were used for testing \((X_{\text{test}})\). This split allows for the evaluation of covariance estimators on unseen data.

\textbf{Z-scoring:} Each filtered time series \(X\) (of shape \((N, D)\), where \(N\) is the number of time points and \(D\) is the number of brain regions) underwent independent Z-score standardization for each dimension (brain region). This ensures that each regional signal has zero mean and unit variance across time, preventing features with larger magnitudes from dominating the covariance estimation. A small epsilon \((10^{-8})\) was added to the standard deviation to avoid division by zero for flat signals:

\begin{equation*}
    X_{\text{zscored}}(t, d) = \frac{X(t, d) - \mu_d}{\sigma_d + \epsilon}
\end{equation*}

where \(\mu_d\) and \(\sigma_d\) are the mean and unbiased standard deviation of region \(d\) over time \(t\), respectively.

%\newpage
%\input{checklist_completed.tex}

\end{document}